%% file: main.tex
\documentclass{article}

\usepackage{arxiv}
\usepackage{amsmath}
\usepackage{array}
\usepackage{booktabs}
\usepackage{caption}
\usepackage{float}
\usepackage{graphicx}
\usepackage{makecell}
\usepackage{multirow}
\usepackage{placeins}
\usepackage{subcaption}
\usepackage[table]{xcolor}
\usepackage[numbers,sort&compress]{natbib}
\usepackage[hyperfootnotes=false]{hyperref}

\definecolor{linkblue}{HTML}{0071BC}
\definecolor{linkred}{HTML}{ED1C24} 
\hypersetup{
  colorlinks=true,
  linkcolor=linkblue,
  citecolor=linkblue,
  urlcolor=black,
}

\definecolor{heroblue}{RGB}{220,230,239} 
\newcommand{\best}[1]{\textbf{#1}}
\newcommand{\second}[1]{\underline{#1}}

\definecolor{deltaup}{RGB}{74,138,101}
\definecolor{deltadown}{RGB}{184,84,80}
\definecolor{deltasame}{RGB}{128,128,128}
\newcommand{\up}[1]{\textcolor{deltaup}{$\uparrow$\,#1\%}}
\newcommand{\down}[1]{\textcolor{deltadown}{$\downarrow$\,#1\%}}
\newcommand{\same}{\textcolor{deltasame}{0.0\%}}

\title{\texorpdfstring{HERO: Histology Encoder for Robust\\Representation in Oncology}{HERO: Histology Encoder for Robust Representation in Oncology}}
\renewcommand{\shorttitle}{HERO: Histology Encoder for Robust Representation in Oncology}
\author{%
  \normalfont
  \parbox{0.9\textwidth}{\centering
    Zhi~Li$^1$, Eghbal~Amidi$^1$, Yating~Cheng$^1$, Tyson~Dawson$^1$, Gorkem~Can~Ates$^1$,
    Shuzhen~Kuang$^1$, Norsang~Lama$^1$, Md~Ashequr~Rahman$^1$, Zhiying~Lu$^1$,
    Elisabeth K. Kong$^1$, Milan~Radovich$^1$, David~Spetzler$^1$,
    George~W.~Sledge$^1$, and Ming~Chen$^{1,\ddagger}$
  }\\[1.4em]
  {\normalfont $^1$Caris Life Sciences, Irving, TX, United States}
}
\date{}

\begin{document}
\maketitle
\begingroup
\renewcommand{\thefootnote}{\ensuremath{\ddagger}}
\footnotetext{Corresponding author. Ming Chen, mchen AT carisls DOT com}
\endgroup
\vspace{-2\baselineskip} 

\begin{abstract}
Foundation models trained on large pathology image corpora now provide strong, transferable representations for computational pathology. Over the past few years a series of such models has been released, each trained on more slides than the last; on standard classification and segmentation benchmarks, the leading models are now separated by small margins. In clinical use, however, the foundation model is applied to images from hospitals, scanners, and staining protocols outside its training data. Encoders generally embed these acquisition factors alongside biological information, which may introduce downstream errors and hinder safe clinical adoption. A pathology foundation model should therefore be robust to acquisition shift without giving up representation quality, yet robustness is seldom the axis along which models are compared. In this report, we introduce HERO (Histology Encoder for Robust Representation in Oncology), a ViT-G/14 pathology foundation model trained with the DINO and iBOT objectives and refined with high-resolution Gram anchoring on a morphology-balanced corpus of 500 million tiles from approximately 575,000 clinical whole-slide images. Across the evaluated public benchmarks, HERO shows the strongest robustness to center, scanner, and stain variation among the compared state-of-the-art foundation models, performs comparably on tile-level classification, segmentation, and gene-expression prediction, ranks first on average across 39 evaluated slide-level clinical tasks, and, under an equal-weighted framework-level analysis, has the best average rank across the six benchmark frameworks.
\end{abstract}

\begin{figure}[H]
\centering
\includegraphics[width=0.86\linewidth]{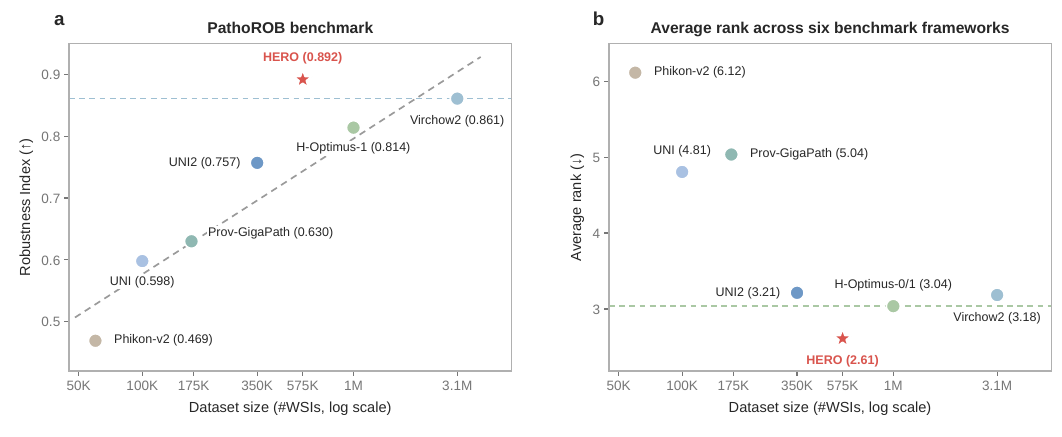}
\caption{\textbf{Robustness and overall standing versus pretraining dataset size.} (a) PathoROB Robustness Index (higher is better). The gray dashed line is a log-linear fit to the six baselines. (b) Average rank across the six benchmark frameworks, each weighted equally (lower is better). Colored dashed lines mark the strongest public baseline. Per-framework ranks are in Table~\ref{tab:overall_ranking}.}
\label{fig:overall_ranking}
\end{figure}

\input{sections/01_introduction}
\input{sections/02_related_work}
\input{sections/03_data_methods}
\input{sections/04_results}
\FloatBarrier
\input{sections/05_discussion}

\begingroup
\small
\bibliographystyle{plainnat-trunc} 
\bibliography{references}
\endgroup

\newpage
\appendix
\input{sections/06_appendix}

\end{document}

%% file: sections/01_introduction.tex
\section{Introduction}

Hematoxylin-and-eosin (H\&E) whole-slide images (WSIs) are collected for nearly every cancer patient, and their digitization has made computational pathology a routine tool rather than a research exercise. Deep learning on these images has already supported diagnosis in clinical validation studies \citep{raciti2023clinical} and predicts biomarkers and molecular phenotypes from morphology alone \citep{diao2021human,binder2021morphological}, but each of these systems was trained for one task. Foundation models now sit at the center of this field. Trained by self-supervision on millions of unlabeled tiles, a single vision transformer yields frozen representations that transfer to diagnosis, subtyping and grading, biomarker and mutation prediction, treatment response, and survival \citep{chen2024towards,wang_pathology_2024,biswas2026causal}. UNI \citep{chen2024towards}, Virchow2 \citep{zimmermann2024virchow2}, Prov-GigaPath \citep{Xu2024}, H-Optimus \citep{hoptimus0,hoptimus1}, and Phikon-v2 \citep{filiot_phikon-v2_2024} all perform well across these tasks, and each generation has been larger than the last in slides, tiles, and parameters. Scale has been the dominant design axis \citep{campanella2023computational,kaiko2024towardslarge}.

Real-world use places a different demand on the encoder. A foundation model is trained once and then kept frozen. Its features are then extracted from slides produced by hospitals, scanners, and staining protocols that were not represented in its training data. Increasingly, these features also feed downstream models that pool patient cohorts from many institutions \citep{wang_pathology_2024,biswas2026causal}. Whatever acquisition signal the encoder carries, the downstream model inherits. Recent measurements show that current encoders do carry it: embeddings cluster by medical center \citep{dejong_pathology_fm_robustness,lin2025unveiling,koemen2024batcheffects}, and predictions shift with scanner and stain \citep{carloni2025pathology,chai2025impact,gustafsson2024evaluating}. A high transfer average does not reveal this behavior, because transfer benchmarks are typically evaluated within a fixed acquisition setting. Robustness to acquisition shift therefore has to be measured on its own, and it is the property that real-world downstream use depends on. Recent studies now make this possible. PathoROB tests whether the nearest neighbors of a tile share its biology rather than its center \citep{komen2025pathorob}, whereas PLISM provides the same tissue rescanned and restained under matched conditions \citep{ochi2024plism,filiot2025plismbench}.

In this work, we present HERO, a ViT-G/14 pathology foundation model trained on approximately 575,000 clinical WSIs digitized on three scanner platforms. Tiles are indexed in a frozen feature space and sampled under per-cluster quotas, so the 500 million tile training corpus is balanced by morphology rather than by raw frequency. The backbone is trained with the DINO and iBOT objectives \citep{caron2021dino,zhou2021ibot,oquab2023dinov2} and then refined with Gram anchoring, which preserves the structure of dense patch tokens \citep{simeoni2025dinov3}. We evaluate two questions: whether HERO is robust to acquisition shift and whether it retains broad downstream representation quality. We therefore use six public benchmarks: robustness to acquisition shift on PathoROB and PLISM \citep{komen2025pathorob,ochi2024plism,filiot2025plismbench}; tile-level classification, segmentation, and stress tests on EVA and THUNDER \citep{kaiko2024eva,marza2025thunder}; gene-expression prediction on HEST \citep{jaume2024hest}; and 39 slide-level clinical tasks on Patho-Bench \citep{zhang2025pathobench}. HERO has the highest Robustness Index on PathoROB despite being trained on a far smaller archive than the models closest to it (Fig.~\ref{fig:overall_ranking}a). Averaged over all six benchmarks, it also has the best rank among the leading pathology foundation models (Fig.~\ref{fig:overall_ranking}b).

%% file: sections/02_related_work.tex
\section{Related Work}

\paragraph{Pathology foundation models.} Over the past few years, pathology foundation models have become the standard starting point for clinical-grade applications in digital pathology \citep{wang2022transformer,ciga2022self,filiot2023scaling,chen2024towards,vorontsov2024virchow,zimmermann2024virchow2,Xu2024,dippel2024rudolfv,hoptimus0,hoptimus1,juyal2024pluto,padigela2025pluto4,nechaev2024hibou,alber2025atlas,alber2026atlas2,kaplan2025openmidnight,wang_pathology_2024,dippel2024nejm}. They come in two forms. Tile-based models encode small, fixed-size image tiles into compact embeddings and are trained with self-supervised frameworks, first contrastive and masked-image objectives \citep{wang2022transformer,ciga2022self,filiot2023scaling} and now almost universally DINOv2 or DINOv3 \citep{caron2021dino,zhou2021ibot,oquab2023dinov2,simeoni2025dinov3}. Slide-based models take these tile embeddings as input and learn to aggregate them into one representation per WSI, using slide-level supervision from pathology reports \citep{shaikovski_prism_2024,vorontsov2025prism2unlockingmultimodalgeneral,ding2025titan,xu2025multimodal} or molecular profiles \citep{vaidya2025threads}. Progress in the tile-based family has come from two levers: more WSIs \citep{campanella2023computational,kaiko2024towardslarge,Xu2024,zimmermann2024virchow2,alber2026atlas2} and larger encoders \citep{zimmermann2024virchow2,hoptimus1,padigela2025pluto4,alber2026atlas2}. Both levers now show diminishing returns. Models trained on orders of magnitude fewer slides reach results close to the largest models \citep{KDK_Training_MICCAI2025,kaplan2025openmidnight}, and on standard tile-level benchmarks the leading encoders are separated by small margins.

\paragraph{Robustness of pathology foundation models.} Although each new model is trained on more slides than the last, the performance gained per added slide is shrinking, and there is no guarantee that robustness improves alongside it, because every proprietary archive carries its own distribution of scanners, stains, and centers. Sensitivity to scanner hardware, laboratory processing, and staining variation is still observed in current models \citep{dejong_pathology_fm_robustness,koemen2024batcheffects,lin2025unveiling,carloni2025pathology,chai2025impact,gustafsson2024evaluating,howard2021site}, and magnification is a further axis along which performance shifts \citep{moellers2026gap}. When prediction labels correlate with such confounders, the representation can pick up the confounder in place of the biology, and performance drops once the correlation breaks \citep{lapuschkin2019clever,kauffmann2025clever,komen2025pathorob,howard2021site}. Several benchmarks now measure this behavior directly. PathoROB quantifies whether nearest-neighbor structure reflects biology rather than center \citep{komen2025pathorob}, PLISM tests embedding consistency and retrieval across matched scanner and stain changes of the same tissue \citep{ochi2024plism,filiot2025plismbench}. Work on improving robustness remains limited and mostly acts after pretraining: stain normalization \citep{macenko,reinhard,komen2025pathorob}, stain and metadata-guided augmentation \citep{tellez2019quantifying,shen2022randstainna,drexlin2025medi}, post-hoc correction of the embedding space by batch-effect correction or supervised re-embedding \citep{komen2025pathorob,nguyen2026fmmap}, distillation into smaller encoders \citep{filiot2025plismbench}, and domain-adversarial training of the downstream head \citep{ganin2016dann,komen2025pathorob}. The composition of the pretraining corpus itself has rarely been treated as a lever for robustness.

\paragraph{Data curation for self-supervised learning.} A clinical archive is not a balanced sample of tissue morphology. A small number of frequent patterns, such as stroma, necrosis, and the common carcinomas of high-volume organs, account for most of the extracted tiles, while rare morphologies contribute few. Sampling tiles in proportion to their raw frequency therefore gives common patterns most of the optimization updates, and the visual coverage of the resulting encoder follows the archive rather than the pathology. Curating the pretraining corpus is therefore critical for addressing this issue. In natural images, DINOv2 assembled its training set by retrieval from a larger uncurated pool \citep{oquab2023dinov2}, clustering-based sampling was later formalized to balance visual concepts \citep{vo2024curation}, and DINOv3 scaled this to hierarchical clustering with balanced sampling \citep{simeoni2025dinov3}. Related work removes near-duplicates and prunes easy examples \citep{abbas2023semdedup,sorscher2022beyond} or balances the concept distribution of web-scale image-text data \citep{xu2024metaclip}. In pathology, the same idea has begun to appear. Feature-space curation has been revisited for pathology encoders \citep{chen2025curation}, and a recent report found that variety across WSIs, rather than the number of tiles drawn from each, drives learning in pathology encoders \citep{bosch2026diversity}. HERO follows the first line of work and applies cluster-quota sampling to a clinical archive of approximately 575,000 slides, so that how often each visual pattern enters optimization is set explicitly rather than inherited from slide and tile frequency.

%% file: sections/03_data_methods.tex
\section{Data and Methods}

\subsection{Overview}

HERO is a tile-level pathology vision encoder produced by a three-part workflow: morphology-balanced corpus construction, self-supervised backbone pretraining, and dense-feature refinement (Fig.~\ref{fig:curation_training}). Data curation converts a large clinical WSI archive into a tile corpus whose exposure to recurring visual patterns is set by cluster quotas rather than by raw tile frequency. Pretraining then trains a ViT-G/14 backbone with the DINO and iBOT objectives \citep{caron2021dino,zhou2021ibot}. The second stage adds a Gram-anchoring loss that preserves dense patch-token structure \citep{simeoni2025dinov3}. The remainder of this section describes tiling and preprocessing, the curation procedure, the backbone and Stage~1 pretraining, and the Stage~2 refinement.

\subsection{Dataset and preprocessing}

\begin{figure}[p]
\centering
\includegraphics[width=0.9\linewidth]{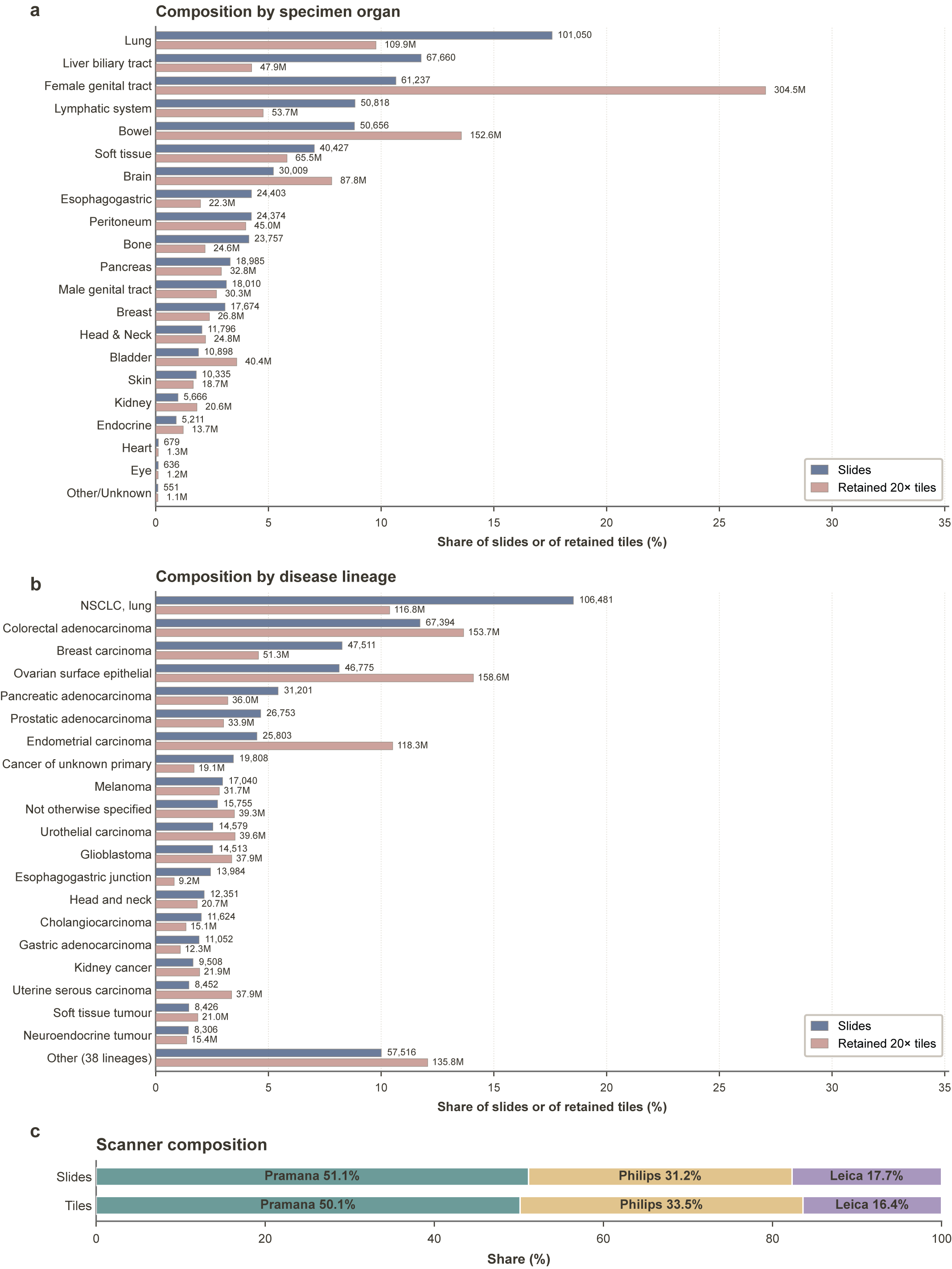}
\caption{\textbf{Composition of the HERO pretraining archive} (approximately 575,000 WSIs; 1.126 billion retained 20$\times$ tiles). (a) Share of slides and of retained tiles by specimen organ, with counts annotated. (b) The same by disease lineage; the 38 smallest lineages are pooled as \emph{Other}. (c) Scanner composition by slide and by tile.}
\label{fig:data_distribution}
\end{figure}

The HERO pretraining archive comprises approximately 575,000 WSIs from 21 organ groups and 58 disease lineages, digitized on three scanner platforms (Fig.~\ref{fig:data_distribution}). We extracted tiles at 0.5~$\mu$m per pixel, which corresponds to 20$\times$ objective magnification. A coarse tissue mask on a downsampled thumbnail first removed near-empty regions, and an HSV threshold then kept foreground tissue. From the retained regions we read non-overlapping $784\times784$ crops at native resolution and Lanczos-resized them to $392\times392$ pixels, giving approximately 1.126 billion tiles. A slide's share of the archive and its share of the tiles can be very different. Female genital tract specimens make up 7\% of the slides but 27\% of the tiles, because these sections are large; lung specimens are 18\% of the slides and 10\% of the tiles (Fig.~\ref{fig:data_distribution}a). The same holds at the lineage level (Fig.~\ref{fig:data_distribution}b). Sampling tiles uniformly would therefore train the encoder mostly on a handful of large, common tissue types.

\subsection{Morphology-balanced curation}

\begin{figure}[t]
\centering
\includegraphics[width=\linewidth]{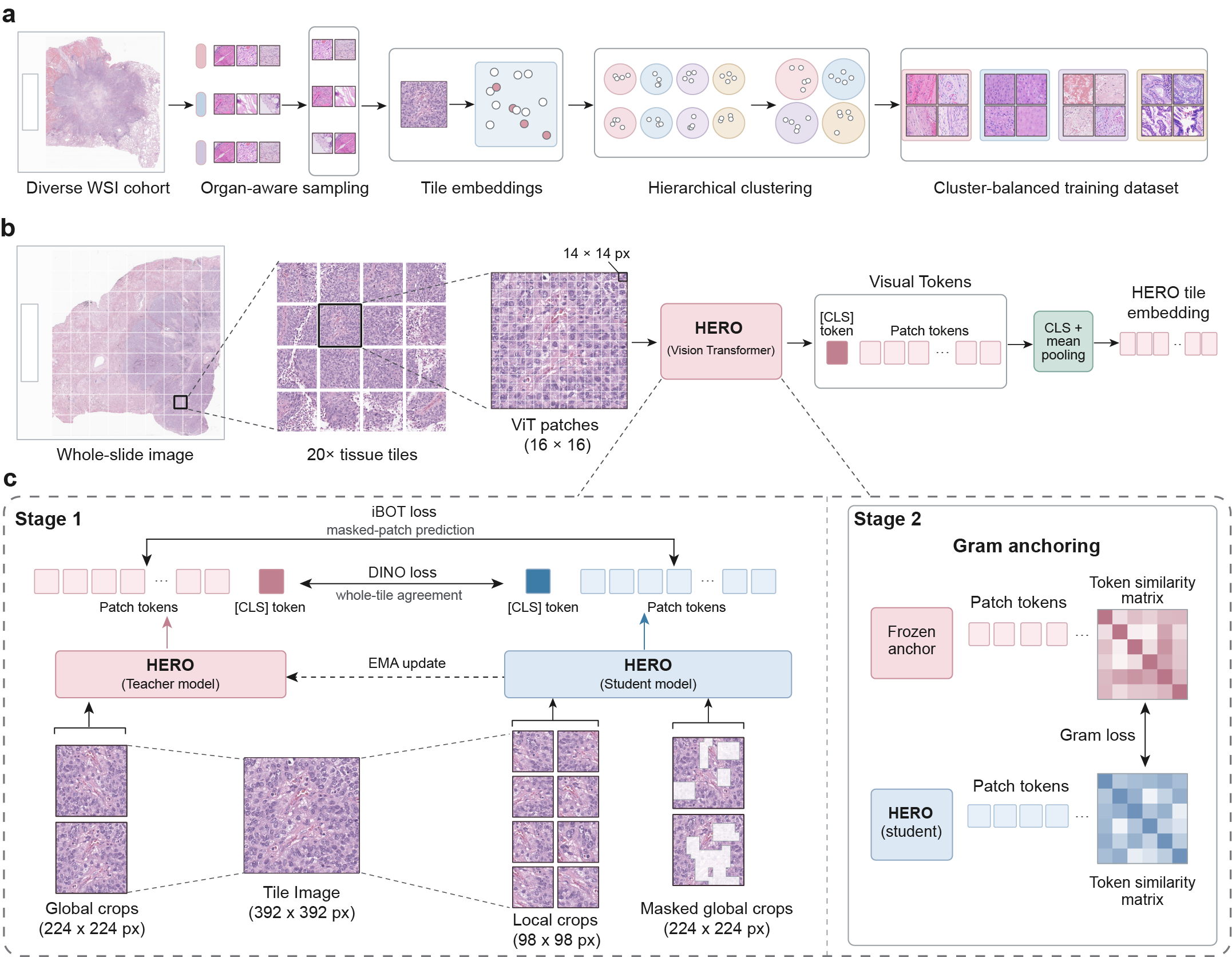}
\caption{\textbf{HERO workflow.} (a) Curation: tiles are embedded with a frozen DINOv2 encoder, clustered at two levels, and sampled under coarse-cluster quotas. (b) Representation extraction: normalized CLS token, mean patch token, and their concatenation (CLS+Mean). (c) Training: Stage~1 uses DINO and iBOT objectives; Stage~2 adds Gram anchoring.}
\label{fig:curation_training}
\end{figure}

Sampling the raw tile pool in proportion to frequency gives common morphologies more optimization updates than rare ones. To control this exposure without diagnostic labels, we index the tile pool in a frozen feature space and sample under per-cluster quotas (Fig.~\ref{fig:curation_training}a).

We first embed each tile with a frozen DINOv2 ViT-B/14 encoder with register tokens \citep{oquab2023dinov2,darcet2024visiontransformersneedregisters}, using $224\times224$ inputs normalized with ImageNet statistics and taking the normalized 768-dimensional CLS representation. Register tokens absorb high-norm artifacts that can appear in dense feature maps \citep{darcet2024visiontransformersneedregisters}.

Curation then proceeds in three steps. First, we fit 100,000 clusters with $k$-means over a 48-million-tile sample allocated across organs in proportion to the square root of retained tile count. This organ-aware allocation is used only to fit the cluster space and does not set the training mixture. Second, we assign the full tile pool to the fine clusters and aggregate the fine-cluster centroids into 10,000 clusters. These clusters partition visual features and are not pathology labels. The two-level index follows clustering-based data curation for self-supervised learning \citep{vo2024curation,chen2025curation}, with cluster counts and quotas specific to this archive.

Third, each cluster $c$ receives a capped quota
\begin{equation}
q_c=\min(n_c,L), \qquad \sum_c q_c=500{,}000{,}000,
\end{equation}
where $n_c$ is the number of tiles in cluster $c$ and $L$ is chosen to meet the corpus target. Small clusters are retained without replacement, and high-frequency clusters are capped, so the final corpus contains 500 million unique tiles with no duplicates. The training distribution is therefore set by visual-cluster quota rather than by raw organ or tile frequency.

\subsection{Backbone and self-supervised pretraining (Stage 1)}

The HERO backbone is a ViT-G/14 with four register tokens and 1,536-dimensional token embeddings \citep{dosovitskiy2020image,darcet2024visiontransformersneedregisters}. It uses a SwiGLU feed-forward layer and stochastic depth with a drop-path rate of 0.4, and has approximately 1.1 billion parameters. Training uses bfloat16 mixed precision with fully sharded data parallelism (FSDP).

Stage~1 trains a student network and an exponential-moving-average teacher with the DINO image-level objective and the iBOT masked patch-level objective \citep{caron2021dino,zhou2021ibot}, following the DINOv2 recipe \citep{oquab2023dinov2}. Each objective uses a separate head with 131,072 prototypes, teacher outputs are centered with Sinkhorn--Knopp normalization, and iBOT masks each image with probability 0.5 at a masking ratio between 0.1 and 0.5. In this stage, we replace the KoLeo regularizer of DINOv2 with the kernel density estimator (KDE) regularizer introduced by Virchow2 \citep{zimmermann2024virchow2}, which spreads embeddings across the unit sphere in the same way but stays stable when many tiles in a batch look alike, a common situation in histopathology. 

Augmentation follows the DINOv2 multi-crop recipe \citep{caron2020swav,oquab2023dinov2}: each tile yields two global crops at $224\times224$ and eight local crops at $98\times98$ by random resized cropping. Every crop then passes through random horizontal and vertical flips, color jitter, grayscale, Gaussian blur, and a stain perturbation in HED space. The latter decomposes the crop into hematoxylin, eosin, and residual channels by color deconvolution \citep{ruifrok2001quantification} and jitters each channel independently \citep{tellez2018whole,tellez2019quantifying}, which exposes the encoder to the range of staining seen across laboratories.

We optimize with AdamW using a linear warmup followed by a constant learning rate, the schedule adopted in DINOv3 to allow training of unbounded length \citep{simeoni2025dinov3}. The learning rate is scaled with the square root of the global batch size relative to a reference batch of 2,048 \citep{malladi2022sdes}. Stage~1 runs for 450k steps on 32 A100 GPUs at a global batch size of 1,536, and the full configuration is listed in Table~\ref{tab:training_config}.

\subsection{Gram-anchored dense refinement (Stage 2)}

DINOv3 reported that long self-supervised training keeps improving the image-level representation while the patch tokens degrade \citep{simeoni2025dinov3}. Their pairwise similarities lose the local structure that segmentation and other dense tasks depend on. They proposed Gram anchoring to counter this drift, and we apply it in the refinement training stage. We resume from the Stage~1 model and train for a further 20k steps with the Gram-anchoring loss added to the DINO and iBOT terms.

The Gram loss compares the token-to-token similarity matrix of the student with that of a frozen anchor network. We use the Stage~1 teacher at 200k steps as the anchor, a checkpoint from before the dense degradation began. The anchor sees each global crop at $448\times448$ pixels without color augmentation, which gives a cleaner and higher resolution similarity target than the student's own view. The loss is normalized and weighted at 0.1 relative to the DINO and iBOT terms. Stage~1 and Stage~2 together take approximately ten days on 32 A100 GPUs. Appendix~\ref{sec:stage2} compares the Stage-1-only model with the released model after Gram anchoring.

\subsection{Evaluation overview}
\label{sec:eval_overview}

We evaluate HERO on six public benchmark frameworks comprising 71 reported metrics, grouped by the property each one measures. PathoROB \citep{komen2025pathorob} and PLISM \citep{ochi2024plism,filiot2025plismbench} measure robustness to center, scanner, and stain variation. EVA \citep{kaiko2024eva} and THUNDER \citep{marza2025thunder} cover tile-level classification and segmentation, and THUNDER adds calibration and adversarial tests. HEST \citep{jaume2024hest} measures gene-expression prediction from tiles. Patho-Bench \citep{zhang2025pathobench} defines 95 slide-level clinical tasks; here, we use the 39 tasks whose datasets we could download under our institution's data-access policy. Task inclusion was determined by data availability and access permissions rather than model performance. We evaluate every compared model on this accessible 39-task subset under the same official protocol; performance across the complete 95-task benchmark may differ. In every benchmark framework, the encoder stays frozen. Tile-level tasks train a linear or small segmentation head on the tile embeddings and slide-level tasks train an attention-based multiple-instance learning aggregator (ABMIL) \citep{ilse2018attention} over the tile embeddings of each slide.

Baseline numbers come from two sources. For PathoROB, PLISM, EVA, THUNDER, and HEST we take the published results of each framework as aggregated by Histoboard, a public leaderboard of pathology foundation models \citep{histoboard2026}, and run only HERO through the same official protocol for each benchmark. Patho-Bench has no such leaderboard for our task set, so for this framework, we reran every comparison model under the same official protocol. Tasks, metrics, and protocols for each framework are listed in Appendix~\ref{sec:eval_protocols}. Appendix~\ref{sec:stage2} presents an ablation study of high-resolution Gram anchoring on EVA, HEST, and PathoROB.
\FloatBarrier

%% file: sections/04_results.tex
\section{Results}

We report the six benchmark frameworks in the order of Section~\ref{sec:eval_overview}: robustness first, then tile-level representation transfer, gene-expression prediction, and slide-level tasks. Tables~\ref{tab:pathorob_results}--\ref{tab:hest_results} share one layout. The comparison models are sorted by average score, from highest to lowest, and HERO appears in a separate blue-shaded row at the bottom of each table. THUNDER (Table~\ref{tab:thunder_results}) is the exception: it reports six metrics with no single average, so its models keep the order of the original leaderboard. In every table, bold marks the best value in a column, and underlining marks the second-best value.

\subsection{Robustness to center, scanner, and stain variation}

To test whether HERO embeddings organize tissue by biology rather than by the center that produced the slide, we evaluated the PathoROB Robustness Index on Camelyon, TCGA, and Tolkach ESCA (Table~\ref{tab:pathorob_results}). The index is the fraction of a tile's nearest neighbors that share its biological class but come from a different center. HERO reached 0.836, 0.884, and 0.956 on the three datasets, for an average of 0.892, and it led on every dataset. Plotted against archive size, it sits above the trend that the six baselines follow (Fig.~\ref{fig:overall_ranking}a). Virchow2 was second with 0.861 and H-Optimus-1 third with 0.814. Camelyon separated the models most: values ranged from 0.019 for Phikon-v2 to 0.836 for HERO, so on this dataset most encoders group tiles by hospital before they group them by metastasis status.

\input{tables/04_pathorob_results}

PLISM asks a narrower question: when the same tissue section is rescanned on a different device or restained with a different protocol, does the embedding remain stable? Table~\ref{tab:plism_results} reports cosine similarity between matched pairs and Top-10 retrieval accuracy across scanner, stain, and combined changes. HERO reached 0.933 in embedding consistency and 0.793, 0.424, and 0.280 in the three retrieval settings, for an average of 0.607. The next best model, H-Optimus-0, averaged 0.480. Stain changes moved every encoder more than scanner changes did, and HERO was the only model above 0.4 on the stain axis.

\input{tables/05_plism_results}

HERO obtained these results with a smaller archive than the comparison models. It was trained on approximately 575,000 WSIs from one laboratory, compared with 3.1 million for Virchow2 and 1 million for H-Optimus-1 (Table~\ref{tab:pathorob_results}). Within this set of models, pretraining slide count alone did not track robustness; this descriptive comparison does not isolate the effects of archive composition, curation, model architecture, or training strategy.

\subsection{Tile-level classification and segmentation}

EVA covers eight tile-level and two slide-level tasks with a linear head for classification, a decoder head for segmentation, and attention-based multiple-instance learning for the slide tasks (Table~\ref{tab:eva_results}). HERO averaged 0.780. UNI2 averaged 0.793 and Virchow2 0.786. HERO had the best Gleason grading score (0.793) and the second-best PCam$_{10}$ score (0.875), and it was within 0.02 of the leader on five of the ten tasks. Its largest deficits were on BACH (0.848 versus 0.915 for UNI2), BRACS, and PANDA, and it trailed H-Optimus-0 on the two nuclear segmentation tasks by 0.011 and 0.028.

\input{tables/02_eva_results}

THUNDER evaluates the same frozen features under six protocols: kNN, linear-probe, and few-shot classification, patch-token segmentation, calibration, and adversarial attack (Table~\ref{tab:thunder_results}). HERO ranked second on linear probing (0.853) and segmentation (0.690), third on kNN (0.825), and fifth on few-shot classification (0.756). On the two remaining protocols it ranked seventh of eight, with an expected calibration error of 4.1\% and an adversarial F1 drop of 44.5\%. Calibration values fell within a relatively narrow range across models. Calibration is computed on the confidence of the linear probe, and the expected calibration error ranged from 3.4\% to 5.8\% across the eight models, so HERO's value is 0.7 points above the best. The adversarial test applies a white-box Projected Gradient Descent attack, five gradient steps on the input pixels constrained to an $\ell_\infty$ ball and using the encoder's own gradients. Under this attack every model lost at least 31\% of its F1. HERO's 44.5\% drop lies between UNI (40.3\%) and H-Optimus-1 (57.4\%). 

\input{tables/06_thunder_results}
\FloatBarrier

\subsection{Gene-expression prediction}

HEST tests whether tile embeddings carry information about local gene expression. For each of nine tissue-specific tasks, a ridge regression predicts the 50 most variable genes from the embedding, and performance is the Pearson correlation between predicted and measured expression (Table~\ref{tab:hest_results}). HERO averaged 0.404. H-Optimus-1, H-Optimus-0, and UNI2 averaged 0.423, 0.415, and 0.414. HERO was second on PRAD (0.378) and ccRCC (0.271) and within 0.02 of the leader on IDC, PAAD, and LUAD. Its lowest relative position was on COAD and READ, the two colorectal tasks, where the H-Optimus models led.

\input{tables/03_hest_results}
\FloatBarrier

\subsection{Slide-level clinical tasks}

\begin{figure}[t]
\centering
\captionsetup[subfigure]{position=top,labelformat=simple,labelsep=none,font=bf,justification=raggedright,singlelinecheck=false,skip=2pt}
\renewcommand\thesubfigure{\alph{subfigure}}
\begin{subfigure}[t]{\linewidth}
\caption{}
\includegraphics[width=\linewidth]{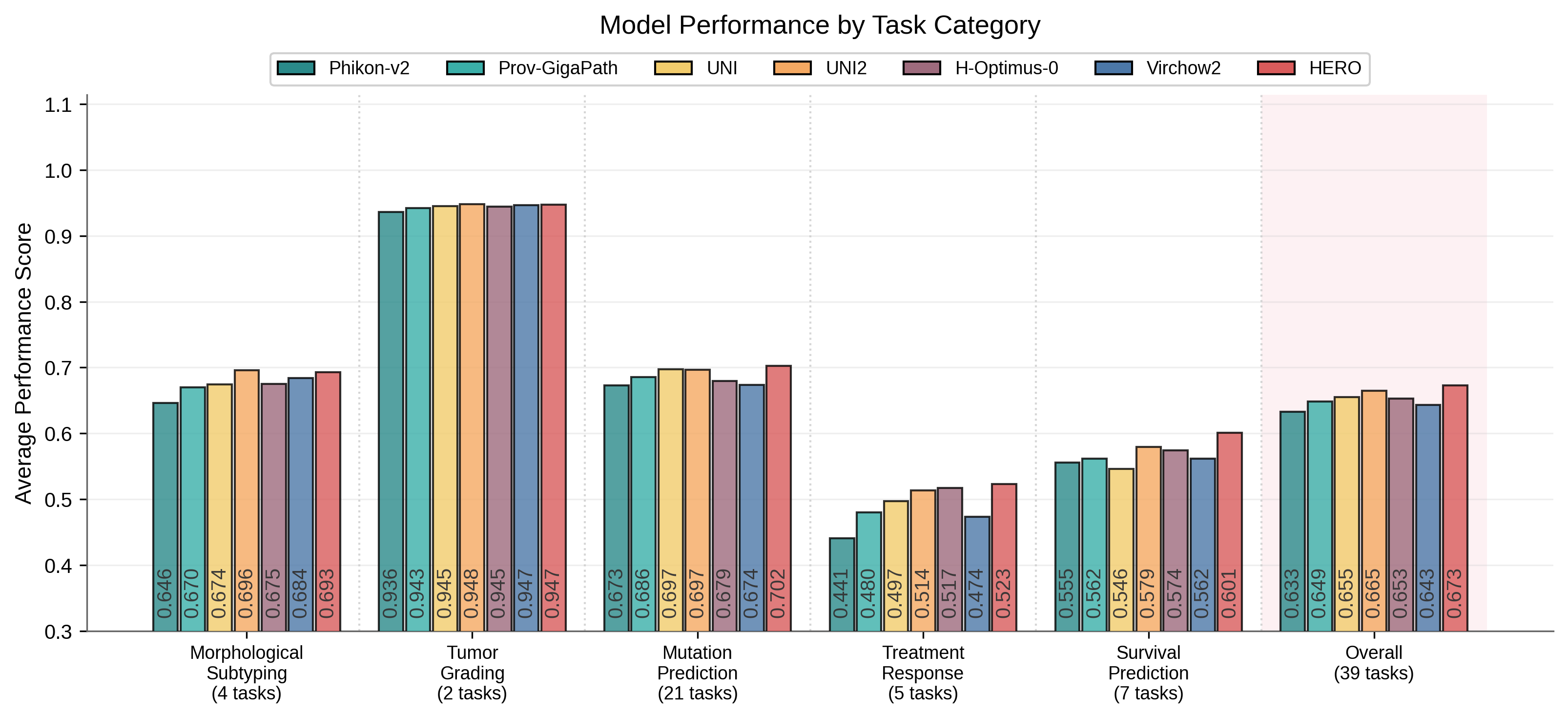}
\end{subfigure}\\[8pt]
\begin{subfigure}[t]{\linewidth}
\caption{}
\includegraphics[width=\linewidth]{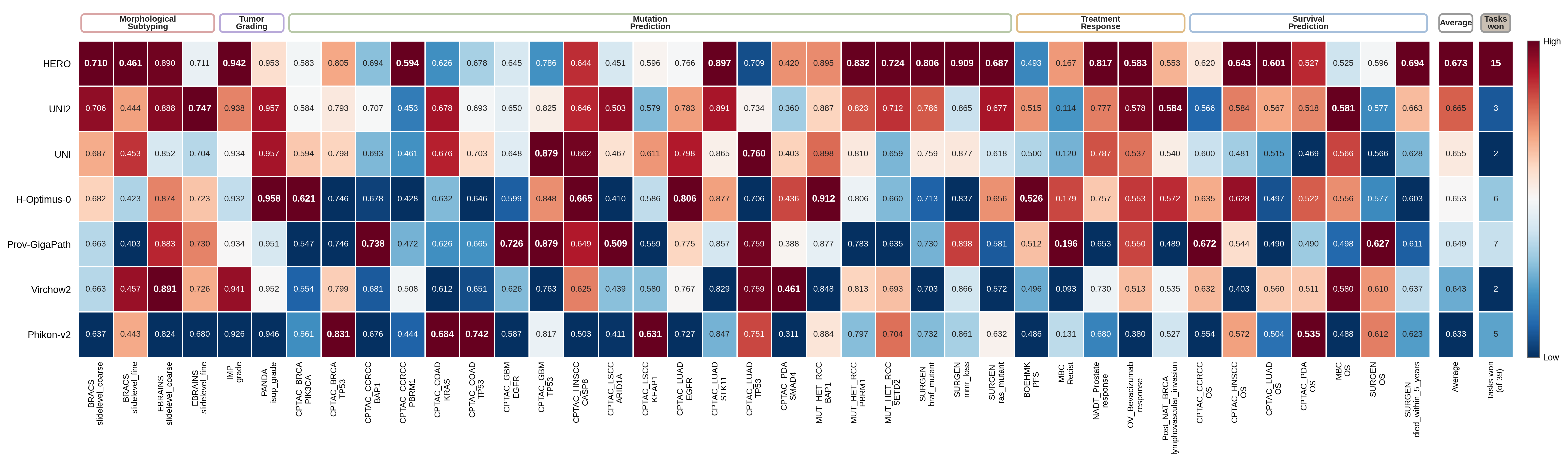}
\end{subfigure}
\caption{\textbf{Slide-level comparison on Patho-Bench (39 tasks).} (a) Category and overall averages for HERO and six public encoders. (b) Per-task scores, color-scaled within each task (blue: low; red: high); best per task in bold. The last two columns give each model's average over the 39 tasks and the number of tasks on which it scored best.}
\label{fig:pathobench}
\end{figure}

Patho-Bench moves from tiles to whole slides. Each of 39 tasks trains an attention-based multiple-instance learning head (ABMIL) on frozen tile embeddings and reports a task-appropriate metric (Figure~\ref{fig:pathobench}). The tasks span morphological subtyping, tumor grading, mutation prediction, treatment response, and survival prediction. HERO averaged 0.673 across all 39 tasks, ahead of UNI2 (0.665), UNI (0.655), H-Optimus-0 (0.653), Prov-GigaPath (0.649), Virchow2 (0.643), and Phikon-v2 (0.633). By category, HERO had the highest average on mutation prediction (0.702), treatment response (0.523), and survival prediction (0.601). UNI2 led morphological subtyping (0.696 versus 0.693) and tumor grading (0.948 versus 0.947), both by less than 0.005. The per-task heatmap further shows that no foundation model led everywhere: HERO scored best on 15 of the 39 tasks, Prov-GigaPath on 7, H-Optimus-0 on 6, and no other model on more than 5. The difference in HERO's overall average was driven mainly by the mutation prediction, treatment response, and survival prediction tasks.

\FloatBarrier

%% file: tables/04_pathorob_results.tex
\begin{table}[H]
\centering
\caption{\textbf{PathoROB results.} Robustness Index (higher means neighbor structure follows biology rather than center). \#WSIs: reported pretraining slide count.}
\label{tab:pathorob_results}
\footnotesize
\setlength{\tabcolsep}{5.5pt}
\renewcommand{\arraystretch}{1.12}
\begin{tabular}{llccccc}
\toprule
Model & \# WSIs & Data Access & \makecell{Camelyon\\{\scriptsize Breast}} & \makecell{TCGA 2$\times$2\\{\scriptsize Multi-organ}} & \makecell{Tolkach ESCA\\{\scriptsize Esophagus}} & \makecell{Average\\{\scriptsize RI $\uparrow$}} \\
\midrule
Virchow2 & 3.1M & Private & \second{0.806} & 0.822 & \second{0.955} & \second{0.861} \\
H-Optimus-1 & 1M & Private & 0.645 & \second{0.853} & 0.944 & 0.814 \\
UNI2 & 350K & Private & 0.544 & 0.803 & 0.923 & 0.757 \\
Prov-GigaPath & 171K & Private & 0.399 & 0.738 & 0.754 & 0.630 \\
UNI & 100K & Private & 0.145 & 0.747 & 0.902 & 0.598 \\
Phikon-v2 & 60K & Public & 0.019 & 0.619 & 0.768 & 0.469 \\
\midrule
\rowcolor{heroblue}\textbf{HERO} & 575K & Private & \best{0.836} & \best{0.884} & \best{0.956} & \best{0.892} \\
\bottomrule
\end{tabular}
\end{table}

%% file: tables/05_plism_results.tex
\begin{table}[H]
\centering
\caption{\textbf{PLISM results.} Median over matched slide pairs for each acquisition axis.}
\label{tab:plism_results}
\footnotesize
\setlength{\tabcolsep}{5.6pt}
\renewcommand{\arraystretch}{1.12}
\begin{tabular}{lccccc}
\toprule
Model & \makecell{Embedding consistency\\{\scriptsize Cosine similarity $\uparrow$}} & \makecell{Scanner\\{\scriptsize Top-10 accuracy $\uparrow$}} & \makecell{Stain\\{\scriptsize Top-10 accuracy $\uparrow$}} & \makecell{Combined\\{\scriptsize Top-10 accuracy $\uparrow$}} & \makecell{Average\\{\scriptsize Metric $\uparrow$}} \\
\midrule
H-Optimus-0 & 0.685 & \second{0.744} & \second{0.327} & \second{0.166} & \second{0.480} \\
Virchow2 & \second{0.777} & 0.609 & 0.306 & 0.163 & 0.464 \\
Prov-GigaPath & 0.570 & 0.592 & 0.118 & 0.054 & 0.333 \\
UNI2-h & 0.591 & 0.501 & 0.190 & 0.046 & 0.332 \\
UNI & 0.547 & 0.532 & 0.169 & 0.053 & 0.325 \\
Phikon-v2 & 0.557 & 0.064 & 0.030 & 0.003 & 0.164 \\
\midrule
\rowcolor{heroblue}\textbf{HERO} & \best{0.933} & \best{0.793} & \best{0.424} & \best{0.280} & \best{0.607} \\
\bottomrule
\end{tabular}
\end{table}

%% file: tables/02_eva_results.tex
\begin{table}[H]
\centering
\caption{\textbf{EVA results.} Balanced accuracy for classification and Dice for segmentation (CoNSeP, MoNuSAC) over eight tile-level and two slide-level tasks.}
\label{tab:eva_results}
\scriptsize
\setlength{\tabcolsep}{2.7pt}
\renewcommand{\arraystretch}{1.12}
\resizebox{\linewidth}{!}{%
\begin{tabular}{lccccccccccc}
\toprule
\multirow{2}{*}{Model}
& \multicolumn{8}{c}{Tile-level} & \multicolumn{2}{c}{Slide-level} & \multirow{2}{*}{Average} \\
\cmidrule(lr){2-9} \cmidrule(lr){10-11}
& \makecell{PCam$_{10}$\\{\scriptsize Bal. acc. $\uparrow$}} & \makecell{BACH\\{\scriptsize Bal. acc. $\uparrow$}} & \makecell{BRACS\\{\scriptsize Bal. acc. $\uparrow$}} & \makecell{CRC\\{\scriptsize Bal. acc. $\uparrow$}} & \makecell{PCam\\{\scriptsize Bal. acc. $\uparrow$}} & \makecell{Gleason\\{\scriptsize Bal. acc. $\uparrow$}} & \makecell{CoNSeP\\{\scriptsize Dice $\uparrow$}} & \makecell{MoNuSAC\\{\scriptsize Dice $\uparrow$}} & \makecell{Cam16\\{\scriptsize Bal. acc. $\uparrow$}} & \makecell{PANDA\\{\scriptsize Bal. acc. $\uparrow$}} & \\
\midrule
UNI2 & \best{0.887} & \best{0.915} & \best{0.661} & \second{0.965} & \best{0.950} & 0.775 & 0.630 & 0.642 & \second{0.849} & 0.657 & \best{0.793} \\
Virchow2 & 0.851 & \second{0.883} & \second{0.624} & \best{0.967} & 0.938 & \second{0.783} & \second{0.640} & 0.669 & \best{0.861} & 0.646 & \second{0.786} \\
H-Optimus-0 & 0.824 & 0.759 & 0.615 & 0.955 & 0.943 & 0.770 & \best{0.644} & \best{0.685} & 0.827 & \best{0.671} & 0.769 \\
Prov-GigaPath & 0.852 & 0.759 & 0.616 & 0.951 & \second{0.945} & 0.724 & 0.626 & \second{0.680} & 0.815 & 0.653 & 0.762 \\
UNI & 0.815 & 0.785 & 0.593 & 0.944 & 0.937 & 0.750 & 0.628 & 0.659 & 0.833 & \second{0.659} & 0.760 \\
Phikon-v2 & 0.820 & 0.729 & 0.568 & 0.940 & 0.920 & 0.729 & 0.627 & 0.635 & 0.798 & 0.644 & 0.741 \\
\midrule
\rowcolor{heroblue}\textbf{HERO} & \second{0.875} & 0.848 & 0.622 & 0.963 & 0.943 & \best{0.793} & 0.633 & 0.657 & 0.834 & 0.636 & 0.780 \\
\bottomrule
\end{tabular}%
}
\end{table}

%% file: tables/06_thunder_results.tex
\begin{table}[H]
\centering
\caption{\textbf{THUNDER results.} Frozen-feature prediction, calibration, and adversarial stress. Adversarial F1 drop is the clean-to-adversarial decrease at the benchmark budget. Parentheses: HERO's within-table rank.}
\label{tab:thunder_results}
\scriptsize
\setlength{\tabcolsep}{3.6pt}
\renewcommand{\arraystretch}{1.12}
\resizebox{\linewidth}{!}{%
\begin{tabular}{lcccccc}
\toprule
Model & \makecell{kNN\\{\scriptsize F1 $\uparrow$}} & \makecell{Linear probe\\{\scriptsize F1 $\uparrow$}} & \makecell{Few-shot\\{\scriptsize F1 $\uparrow$}} & \makecell{Segmentation\\{\scriptsize Dice $\uparrow$}} & \makecell{Calibration\\{\scriptsize ECE (\%) $\downarrow$}} & \makecell{Adversarial\\{\scriptsize F1 drop (\%) $\downarrow$}} \\
\midrule
UNI2 & \best{0.833} & \best{0.857} & \best{0.798} & \second{0.690} & 3.9 & \second{31.7} \\
Virchow2 & \second{0.829} & 0.848 & 0.739 & \best{0.693} & 3.9 & \best{31.1} \\
UNI & 0.808 & 0.835 & \second{0.781} & 0.678 & 3.8 & 40.3 \\
H-Optimus-0 & 0.814 & 0.838 & 0.762 & 0.652 & 4.0 & 43.9 \\
H-Optimus-1 & 0.825 & 0.851 & 0.773 & 0.645 & \second{3.5} & 57.4 \\
Prov-GigaPath & 0.795 & 0.829 & 0.755 & 0.635 & \best{3.4} & 42.1 \\
Phikon-v2 & 0.757 & 0.809 & 0.736 & 0.680 & 5.8 & 33.5 \\
\midrule
\rowcolor{heroblue}\textbf{HERO} & 0.825 {\scriptsize (3)} & \second{0.853 {\scriptsize (2)}} & 0.756 {\scriptsize (5)} & \second{0.690 {\scriptsize (2)}} & 4.1 {\scriptsize (7)} & 44.5 {\scriptsize (7)} \\
\bottomrule
\end{tabular}%
}
\end{table}

%% file: tables/03_hest_results.tex
\begin{table}[H]
\centering
\caption{\textbf{HEST results.} Pearson correlation for the 50 most variable genes in each task.}
\label{tab:hest_results}
\scriptsize
\setlength{\tabcolsep}{3.0pt}
\renewcommand{\arraystretch}{1.12}
\resizebox{\linewidth}{!}{%
\begin{tabular}{lcccccccccc}
\toprule
Model & \makecell{IDC\\{\scriptsize Breast}} & \makecell{PRAD\\{\scriptsize Prostate}} & \makecell{PAAD\\{\scriptsize Pancreas}} & \makecell{SKCM\\{\scriptsize Skin}} & \makecell{COAD\\{\scriptsize Colon}} & \makecell{READ\\{\scriptsize Rectum}} & \makecell{ccRCC\\{\scriptsize Kidney}} & \makecell{LUAD\\{\scriptsize Lung}} & \makecell{LYMPH-IDC\\{\scriptsize Lymph node}} & \makecell{Average\\{\scriptsize Pearson $r\,\uparrow$}} \\
\midrule
H-Optimus-1 & \best{0.602} & \second{0.378} & \second{0.496} & \second{0.659} & \best{0.320} & \best{0.242} & 0.253 & \best{0.578} & \best{0.277} & \best{0.423} \\
H-Optimus-0 & \second{0.598} & \best{0.385} & 0.491 & 0.645 & \second{0.309} & \second{0.222} & 0.268 & 0.559 & 0.259 & \second{0.415} \\
UNI2 & 0.590 & 0.357 & \best{0.500} & \best{0.661} & 0.301 & \second{0.222} & 0.264 & 0.559 & \second{0.273} & 0.414 \\
Virchow2 & 0.597 & 0.353 & 0.478 & 0.640 & 0.258 & 0.207 & \best{0.272} & \second{0.569} & 0.257 & 0.403 \\
Prov-GigaPath & 0.551 & 0.370 & 0.475 & 0.562 & 0.299 & 0.196 & 0.243 & 0.541 & 0.250 & 0.387 \\
UNI & 0.589 & 0.294 & 0.481 & 0.635 & 0.261 & 0.184 & 0.240 & 0.546 & 0.256 & 0.387 \\
Phikon-v2 & 0.533 & 0.342 & 0.443 & 0.535 & 0.262 & 0.153 & 0.242 & 0.547 & 0.237 & 0.366 \\
\midrule
\rowcolor{heroblue}\textbf{HERO} & 0.585 & \second{0.378} & 0.489 & 0.614 & 0.261 & 0.209 & \second{0.271} & 0.561 & 0.269 & 0.404 \\
\bottomrule
\end{tabular}%
}
\end{table}

%% file: sections/05_discussion.tex
\section{Discussion}

A pathology foundation model intended for clinical applications must do two things at once. It must perform well on the task at hand and maintain that performance on slides from hospitals, scanners, or staining protocols not encountered during training---a common situation in real-world precision oncology. Current pathology foundation models are compared almost entirely on the first requirement, and on standard benchmarks the leading models are now separated by small margins. We developed HERO to meet both requirements, with a morphology-balanced corpus from approximately 575,000 clinical slides and a two-stage self-supervised training recipe, and evaluated it on six public benchmark frameworks that measure the two requirements separately. HERO leads both robustness benchmarks, remains competitive with the strongest encoders on tile-level transfer, ranks first on average across the 39 slide-level clinical tasks, and, under the equal-weighted framework-level analysis, has the best average rank across all six benchmark frameworks (Fig.~\ref{fig:overall_ranking}b).

The number of slides in the training dataset alone does not guarantee robustness. Our comparison places HERO above that trend line. Among the six public model baselines, the PathoROB Robustness Index rises log-linearly with archive size (Fig.~\ref{fig:overall_ranking}a). HERO was trained on approximately 575,000 slides, a fraction of the 3.1 million behind Virchow2 and the 1 million behind H-Optimus-1, yet it sits above that line and led both robustness benchmarks (Tables~\ref{tab:pathorob_results} and~\ref{tab:plism_results}). One possible explanation is that each model learns from an archive whose mix of scanners, stains, and centers defines the distribution the encoder treats as normal. We hypothesize that HERO's robustness, and possibly its lead on the slide-level Patho-Bench clinical tasks, reflects three choices acting together: a diverse clinical archive; a cluster-quota curation strategy that caps frequent tissue patterns, allowing rare morphologies and less common acquisition conditions to receive comparable exposure during training; and a stronger self-supervised training recipe.

The ablation in Appendix~\ref{sec:stage2} isolates the last of these. Adding Gram anchoring on top of the Stage~1 model raised the PathoROB Robustness Index from 0.881 to 0.892, with the largest gain on Camelyon, the dataset on which the compared models spread the widest. It did this without reducing overall EVA performance: the average increased from 0.776 to 0.780, with the nuclear segmentation and grading tasks gaining most. The trade-off was a small decrease on the HEST gene-expression prediction benchmark, from 0.414 to 0.404. A 20k-step Gram anchoring refinement that improves the robustness index shows that the training strategy can still influence pathology foundation model performance. We did not independently ablate archive composition or morphology-balanced sampling because of computational constraints and the prioritization of real-world clinical deployment tasks.

On the 39 Patho-Bench clinical tasks, HERO had the best overall average (Fig.~\ref{fig:pathobench}a). UNI2 was ahead on morphological subtyping and tumor grading, by less than 0.005, while HERO led the three larger categories, mutation prediction, treatment response, and survival, and these carried the average. The labels in those three categories are least directly observable by a pathologist, and the signal is subtle and spread across many tiles rather than concentrated in a few diagnostic regions. We hypothesize that this is where data curation provides the greatest benefit. Cluster-quota sampling exposes the encoder to rare morphologies that would be underrepresented by raw-frequency sampling, so the tile embeddings carry information about uncommon patterns. Patho-Bench then trains the ABMIL head with the same default hyperparameters for every encoder. An encoder whose embeddings already separate the rare, informative tiles from the common background gives the aggregator a more tractable problem, which may explain HERO's advantage on the molecular and outcome tasks. The per-task results limit this interpretation: no encoder led on every task (Fig.~\ref{fig:pathobench}b), and broader evaluation on real-world tasks will require contribution from across the field. Interestingly, THUNDER yielded mixed results. HERO ranked second on linear probing and segmentation and third on kNN, but seventh of eight on calibration and on the adversarial test. The two stress metrics measure something other than acquisition robustness. Calibration is the expected calibration error of the linear probe, so it reflects the head as much as the features, and the eight models sit within 2.4 points of one another. The adversarial test perturbs input pixels with a white-box gradient attack that requires access to the model weights and produces imperceptible perturbations; every model lost at least 31\% of its F1 under it. In reality, such perturbations do not represent variation produced by scanners, stains, or laboratories. The adversarial drop describes the loss landscape around each image, and should not be interpreted as predicting behavior under the acquisition shifts that PathoROB and PLISM measure. 

The field's growing focus on robustness is important. If a pathology foundation model stores center, scanner, or stain information in its embeddings, any downstream model inherits it, and a change of scanner or staining protocol can cause predictions to reflect acquisition conditions rather than biology. That sensitivity can make a model fragile in deployment. The evaluations available today measure this at the tile level. What is still missing is slide-level robustness, measured on real prognostic and predictive tasks across institutions, where the aggregator and the encoder either fail or hold up together. Moreover, foundation models are most useful where they see what a pathologist cannot, as in gene expression and mutation prediction from H\&E. For such predictions to be trusted, the models need interpretability that a researcher or physician can act on: which tissue regions drove a prediction, and whether those regions make biological sense. Closing that loop, from model output back to human understanding and forward again to better labels and better models, is where the next gains in clinical value are likely to come from.

%% file: sections/06_appendix.tex
\section{Appendix}

\subsection{Training configuration}

Table~\ref{tab:training_config} lists every setting used in the two training stages. The values are taken from the training configuration files of the reported run.

\begin{table}[H]
\centering
\caption{\textbf{HERO training configuration.} Upper block: settings shared by both stages. Lower block: stage-specific settings; a dash marks settings not used in Stage~1.}
\label{tab:training_config}
\small
\setlength{\tabcolsep}{9pt}
\renewcommand{\arraystretch}{1.2}
\begin{tabular}{@{}l l l@{}}
\toprule
\multicolumn{3}{@{}l}{\textbf{Shared across both stages}} \\
\midrule
Backbone & \multicolumn{2}{l}{ViT-G/14, 4 register tokens, 1{,}536-d, SwiGLU} \\
Parameters & \multicolumn{2}{l}{$\approx$1.1B} \\
Precision & \multicolumn{2}{l}{bfloat16, FSDP} \\
Drop-path rate & \multicolumn{2}{l}{0.4} \\
DINO / iBOT prototypes & \multicolumn{2}{l}{131{,}072 / 131{,}072} \\
KDE weight & \multicolumn{2}{l}{0.05} \\
Masking prob.\ / ratio & \multicolumn{2}{l}{0.5 / 0.1--0.5} \\
Teacher centering & \multicolumn{2}{l}{Sinkhorn--Knopp} \\
Global crops & \multicolumn{2}{l}{$2\times224^2$, scale 0.32--1.0} \\
Local crops & \multicolumn{2}{l}{$8\times98^2$, scale 0.05--0.32} \\
Augmentation & \multicolumn{2}{l}{HED stain jitter, color jitter, grayscale, Gaussian blur, h/v-flip} \\
Optimizer & \multicolumn{2}{l}{AdamW ($\beta_2=0.99$)} \\
LR schedule & \multicolumn{2}{l}{warmup then flat} \\
LR batch scaling & \multicolumn{2}{l}{square-root to batch 2{,}048} \\
Weight decay & \multicolumn{2}{l}{0.04 (constant)} \\
Gradient clip & \multicolumn{2}{l}{3.0} \\
Batch size & \multicolumn{2}{l}{1{,}536 (48 per GPU $\times$ 32)} \\
Hardware & \multicolumn{2}{l}{$32\times$ A100} \\
Wall-clock & \multicolumn{2}{l}{$\approx$10 days (both stages combined)} \\
\midrule
\multicolumn{3}{@{}l}{\textbf{Stage-specific}} \\
\addlinespace[2pt]
 & Stage~1 (pretraining) & Stage~2 (Gram refinement) \\
\cmidrule(lr){2-2}\cmidrule(l){3-3}
Objective & DINO + iBOT & DINO + iBOT + Gram \\
Base learning rate & $2\times10^{-4}$ & $3\times10^{-5}$ \\
Teacher EMA momentum & 0.994 & 0.999 \\
Training steps & 450k & 20k \\
Gram loss weight & -- & 0.1 \\
Gram anchor & -- & Stage~1 teacher @ 200k steps \\
Gram anchor crop & -- & $448^2$, no color distortion \\
\bottomrule
\end{tabular}
\end{table}

\subsection{Evaluation frameworks and benchmark protocols}
\label{sec:eval_protocols}

This section describes each benchmark framework in the order used in the main text: what it measures, which datasets it draws on, how the frozen encoder is evaluated, and which metric we report. For the five benchmark frameworks with a public leaderboard we used the official code and default settings so that HERO's numbers are directly comparable with the Histoboard entries of the other models \citep{histoboard2026}. Table~\ref{tab:benchmark_overview} summarizes the six benchmark frameworks.

\begin{table}[H]
\centering
\caption{\textbf{Evaluation benchmarks.} Datasets counts the constituent datasets or cohorts. Higher is better for all metrics except the two THUNDER stress metrics.}
\label{tab:benchmark_overview}
\small
\setlength{\tabcolsep}{5pt}
\renewcommand{\arraystretch}{1.12}
\begin{tabular}{llclll}
\toprule
Framework & Level & Datasets & Property measured & Protocol & Metric \\
\midrule
PathoROB & Tile & 3 & Center robustness & kNN neighborhood & Robustness Index \\
PLISM & Tile & 1$^a$ & Scanner and stain robustness & Retrieval & Cosine sim., Top-10 acc. \\
EVA & Tile + slide & 10$^b$ & Classification, segmentation & Linear probe, ABMIL & Bal.\ acc., Dice \\
THUNDER & Tile & 20 & Classification, segmentation, stress & kNN, linear probe, few-shot & F1, Dice, ECE, F1 drop \\
HEST & Tile & 9 & Gene-expression prediction & Ridge regression & Pearson $r$ \\
Patho-Bench & Slide & 19$^c$ & Clinical slide-level tasks & ABMIL & AUROC, Bal.\ acc., C-index \\
\bottomrule
\multicolumn{6}{@{}l}{\footnotesize $^a$ One dataset of registered slide pairs (13 stains $\times$ 7 scanners). $^b$ Eight tile-level and two slide-level datasets.} \\
\multicolumn{6}{@{}l}{\footnotesize $^c$ 19 of the 33 Patho-Bench cohorts; the remainder were not available for download under our data-access policy.} \\
\end{tabular}
\end{table}

\paragraph{PathoROB.} PathoROB measures whether a foundation model's embedding space is organized by biology or by the medical center that produced the slide \citep{komen2025pathorob}. Center differences arise from staining, scanner hardware, surgical technique, and laboratory protocol, and are non-biological by construction. The Robustness Index is computed on the $k$ nearest neighbors of each tile, with $k$ fixed per dataset by the benchmark (11 for Camelyon, 61 for TCGA, 46 for Tolkach ESCA), as $\mathcal{R}=|SO|/(|SO|+|OS|)$, where $SO$ counts neighbors that share the tile's biological class but come from another center and $OS$ counts neighbors from the same center with another class. A value of 1 means neighbors are chosen by biology alone. The full resource holds 99,392 tiles from 28 biological classes and 34 centers; the leaderboard uses three balanced multi-center datasets in which every center contributes equally per class. Camelyon, from CAMELYON16 and CAMELYON17 \citep{bejnordi2017diagnostic,bandi2019camelyon17}, is breast lymph-node metastasis detection across two centers. TCGA $2\times2$, built from TCGA-UT \citep{komura2022universal}, comprises 94 quartets of two cancer types from two centers. Tolkach ESCA \citep{tolkach2023esca} is six-class tissue-compartment classification of esophageal adenocarcinoma across three centers. We ran the official implementation with default settings and report the per-dataset index and its mean.

\paragraph{PLISM.} The PLISM dataset registers the same tissue sections after 13 H\&E staining protocols and 7 scanners, covering 46 tissue types, so that any two tiles in a matched pair differ only in acquisition \citep{ochi2024plism}. The PLISM benchmark asks how well an encoder recognizes that they are the same tissue \citep{filiot2025plismbench}. Embedding consistency is the median cosine similarity between the embeddings of a matched pair. Retrieval is Top-10 accuracy: for each tile, the encoder's embedding is used to retrieve the ten nearest tiles under the other acquisition condition, and the retrieval counts as correct if the matched tile is among them. Retrieval is reported separately for cross-scanner pairs (same stain), cross-stain pairs (same scanner), and pairs that differ in both. 

\paragraph{EVA.} EVA is an open evaluation framework for pathology foundation models with fixed data splits and frozen-encoder protocols \citep{kaiko2024eva}. We report the eight tile-level and two slide-level tasks of its default suite. The tile tasks are BACH, four-class breast biopsy classification \citep{aresta2019bach}; BRACS, seven-class breast lesion subtyping \citep{brancati2022bracs}; CRC-100k, nine-class colorectal tissue classification \citep{kather2018crc100k}; PCam, binary lymph-node metastasis detection on patches from CAMELYON16 \citep{veeling2018rotation}, evaluated both with the full training set (PCam) and with ten labeled tiles per class (PCam$_{10}$), following the OpenMidnight evaluation \citep{kaplan2025openmidnight}; Gleason, four-class Gleason pattern grading on prostate tissue microarrays \citep{arvaniti2018gleason}; and two nuclear segmentation tasks, CoNSeP in colorectal tissue \citep{graham2019hovernet} and MoNuSAC across four organs \citep{verma2021monusac}. The slide tasks are Camelyon16, slide-level metastasis detection \citep{bejnordi2017diagnostic}, and PANDA, six-class ISUP grading of prostate biopsies \citep{bulten2022panda}, both in EVA's reduced-sample setting. Classification tasks train a linear head on frozen embeddings, segmentation tasks train a decoder on frozen patch tokens, and slide tasks train an attention-based multiple-instance learning head (ABMIL) \citep{ilse2018attention}. We report balanced accuracy for classification and Dice without background for segmentation.

\paragraph{THUNDER.} THUNDER evaluates frozen tile-level encoders on public tile datasets spanning breast, colorectal, kidney, esophagus, skin, thorax, and multi-organ tissue at magnifications from 5$\times$ to 40$\times$ \citep{marza2025thunder}. The paper describes 16 datasets; the release we ran adds the four SPIDER datasets, for 20 in total, 16 for classification and 4 for segmentation. The classification datasets are BACH \citep{aresta2019bach}, BRACS \citep{brancati2022bracs}, BreakHis \citep{spanhol2016breakhis}, Camelyon17-WILDS \citep{koh2021wilds}, CCRCC \citep{brummer2023ccrcc}, CRC-100k \citep{kather2018crc100k}, ESCA \citep{tolkach2023esca}, MHIST \citep{wei2021mhist}, PCam \citep{veeling2018rotation}, the four SPIDER datasets for breast, colorectal, skin, and thorax \citep{nechaev2025spider}, TCGA CRC-MSI \citep{kather2019msi}, TCGA-TILs \citep{saltz2018tils}, and TCGA Uniform \citep{komura2022universal}. The segmentation datasets are Ocelot \citep{ryu2023ocelot}, PanNuke \citep{gamper2019pannuke,gamper2020pannuke}, and the epithelial and lymphocyte subsets of SegPath \citep{komura2023segpath}. We report six of THUNDER's protocols, all applied to the same frozen features. KNN classification, linear probing, and few-shot classification report macro F1; segmentation trains a head on patch tokens and reports Dice. Calibration reports the expected calibration error of the linear probe. The adversarial test applies a Projected Gradient Descent attack in the $\ell_\infty$ ball to the input pixels, five steps with the encoder's own gradients, and reports the drop in F1 from clean to attacked test images at the benchmark's perturbation budget. Each metric in Table~\ref{tab:thunder_results} is the mean over datasets.

\paragraph{HEST.} HEST-1k pairs spatial transcriptomics profiles with H\&E images, and the HEST-Benchmark turns a subset of 70 samples from 45 patients into gene-expression prediction tasks \citep{jaume2024hest}. For each spatial spot, the encoder embeds the $112 \times 112~\mu$m tile centered on the spot, and a ridge regression on the embeddings reduced to 256 principal components predicts the expression of the 50 most variable genes. We report the nine tissue-specific tasks in Table~\ref{tab:hest_results}: invasive ductal carcinoma (IDC), prostate adenocarcinoma (PRAD), pancreatic adenocarcinoma (PAAD), skin cutaneous melanoma (SKCM), colon adenocarcinoma (COAD), rectal adenocarcinoma (READ), clear cell renal cell carcinoma (ccRCC), lung adenocarcinoma (LUAD), and axillary lymph nodes in IDC (LYMPH-IDC). Splits are patient-stratified, with one fold per patient (two patients per fold for ccRCC). The metric is the Pearson correlation between predicted and measured expression, averaged over genes and folds.

\paragraph{Patho-Bench.} Patho-Bench standardizes slide-level evaluation of pathology encoders \citep{zhang2025pathobench}. Its public release defines 95 tasks over 33 public cohorts in seven categories. We use the 39 tasks whose cohorts we could download, drawn from 19 cohorts, and they cover five of the seven categories (Table~\ref{tab:pathobench_tasks}). Morphological subtyping uses BRACS \citep{brancati2022bracs} and EBRAINS \citep{roetzer2022ebrains}, each at a coarse and a fine label granularity. Tumor grading uses IMP-CRS \citep{neto2024impcrs} and PANDA \citep{bulten2022panda}. Mutation prediction covers 15 gene-level tasks on eight CPTAC cohorts \citep{edwards2015cptac}, three on the MUT-HET-RCC cohort \citep{acosta2022muthetrcc}, and three on SURGEN \citep{myles2025surgen}. Treatment response covers progression-free survival in BOEHMK ovarian cancer \citep{boehm2022boehmk}, RECIST response in the MBC cohort \citep{bergstrom2024mbc}, response to neoadjuvant androgen deprivation in NADT-Prostate \citep{wilkinson2021nadt}, bevacizumab response in OV-Bevacizumab \citep{wang2022ovbev}, and lymphovascular invasion after neoadjuvant therapy in Post-NAT-BRCA \citep{martel2019postnat}. Survival prediction covers overall survival in four CPTAC cohorts, MBC, and SURGEN, and five-year mortality in SURGEN. Every task follows the protocol of the Patho-Bench paper: an attention-based multiple-instance learning head (ABMIL) \citep{ilse2018attention} is trained on the frozen tile embeddings of each slide with the task's objective from the Patho-Bench implementation, using its predefined patient-stratified folds. Metrics follow the benchmark: AUROC for binary classification, balanced accuracy for multi-class classification, quadratic weighted Cohen's kappa for grading, and the concordance index for survival. We ran all seven encoders under this protocol.

\begin{table}[H]
\centering
\caption{\textbf{The 39 Patho-Bench tasks used in Fig.~\ref{fig:pathobench}.} Cohort and task names follow the Patho-Bench naming.}
\label{tab:pathobench_tasks}
\scriptsize
\setlength{\tabcolsep}{4pt}
\renewcommand{\arraystretch}{1.1}
\begin{tabular}{@{}l l l@{}}
\toprule
Category & Cohort & Tasks \\
\midrule
Morphological subtyping (4) & BRACS & slide-level coarse; slide-level fine \\
 & EBRAINS & slide-level coarse; slide-level fine \\
\addlinespace[2pt]
Tumor grading (2) & IMP-CRS & grade \\
 & PANDA & ISUP grade \\
\addlinespace[2pt]
Mutation prediction (21) & CPTAC-BRCA & PIK3CA; TP53 \\
 & CPTAC-CCRCC & BAP1; PBRM1 \\
 & CPTAC-COAD & KRAS; TP53 \\
 & CPTAC-GBM & EGFR; TP53 \\
 & CPTAC-HNSCC & CASP8 \\
 & CPTAC-LSCC & ARID1A; KEAP1 \\
 & CPTAC-LUAD & EGFR; STK11; TP53 \\
 & CPTAC-PDA & SMAD4 \\
 & MUT-HET-RCC & BAP1; PBRM1; SETD2 \\
 & SURGEN & BRAF mutant; MMR loss; RAS mutant \\
\addlinespace[2pt]
Treatment response (5) & BOEHMK & progression-free survival \\
 & MBC & RECIST response \\
 & NADT-Prostate & response \\
 & OV-Bevacizumab & response \\
 & Post-NAT-BRCA & lymphovascular invasion \\
\addlinespace[2pt]
Survival prediction (7) & CPTAC-CCRCC; CPTAC-HNSCC; CPTAC-LUAD; CPTAC-PDA & overall survival \\
 & MBC & overall survival \\
 & SURGEN & overall survival; died within 5 years \\
\bottomrule
\end{tabular}
\end{table}

\subsection{Ablation study of high-resolution Gram anchoring in Stage~2}
\label{sec:stage2}

Stage~2 is a short refinement of a finished Stage~1 model, so its effect is a redistribution of performance rather than a uniform gain. We compare the Stage-1-only model (DINO and iBOT, 450k steps) with the released model, which continues it for 20k steps with Gram anchoring (+ Gram anchoring in the tables), on PathoROB, EVA, and HEST. Both models use the same frozen-encoder protocols as the main text and the CLS+Mean readout, so the only difference between the two rows of each table is Stage~2. Every other setting, including the training corpus, follows Table~\ref{tab:training_config}.

Table~\ref{tab:ablation_summary} gives the three benchmark averages. Gram anchoring raised the PathoROB Robustness Index from 0.881 to 0.892 and the EVA average from 0.776 to 0.780, and lowered the HEST average from 0.414 to 0.404.

\input{tables/07_ablation_summary}

On PathoROB (Table~\ref{tab:ablation_pathorob}), Stage~2 improved the Robustness Index on Camelyon ($+2.8\%$) and TCGA $2\times2$ ($+1.3\%$) and left Tolkach ESCA unchanged at 0.956, for a $+1.2\%$ gain in the average. Camelyon, the dataset on which the compared models spread the widest (Table~\ref{tab:pathorob_results}), is where Gram anchoring helped most

\input{tables/10_ablation_pathorob}

On EVA (Table~\ref{tab:ablation_eva}), the two segmentation tasks are the natural place to look first, since DINOv3 reported that Gram anchoring mainly helps dense prediction \citep{simeoni2025dinov3}. MoNuSAC improved by $2.0\%$ and CoNSeP did not change, so the dense-token benefit appeared on one of the two. Among the classification tasks, Gleason grading ($+2.2\%$), BACH ($+1.9\%$), and PCam$_{10}$ ($+1.0\%$) gained, while BRACS ($-1.1\%$), Cam16 ($-0.5\%$), CRC ($-0.4\%$), and PCam ($-0.2\%$) slipped. The tasks that slipped were already near the top of their range.

\input{tables/08_ablation_eva}

On HEST (Table~\ref{tab:ablation_hest}), the direction is mixed and the average moves against Stage~2. LYMPH-IDC ($+3.1\%$), PAAD ($+2.5\%$), IDC ($+2.3\%$), and PRAD ($+1.1\%$) improved, while COAD ($-13.9\%$), SKCM ($-7.8\%$), and ccRCC ($-7.5\%$) fell enough to lower the average by $2.4\%$. We report this trade-off rather than pick a checkpoint per benchmark. Stage~2 was adopted for its dense-token and robustness behavior, and gene-expression regression is where it costs the most.

\input{tables/09_ablation_hest}

These comparisons cover one archive and one Stage-2 configuration, so they describe the adopted recipe rather than the optimal amount of dense refinement. PLISM and THUNDER were not evaluated on the Stage-1-only model, so the robustness evidence for Stage~2 rests on PathoROB.

\subsection{Average rank across benchmark frameworks}

Table~\ref{tab:overall_ranking} lists the average ranks plotted in Fig.~\ref{fig:overall_ranking}b. Within each benchmark framework, models are ranked on every reported metric (1 = best; ties share the average rank): the columns of Tables~\ref{tab:pathorob_results}--\ref{tab:hest_results} and the 39 Patho-Bench tasks, 71 metrics in total. THUNDER therefore enters with its six aggregate metrics rather than its 20 datasets, and PLISM with its four metrics on one dataset. The ranks are averaged within each framework, and the final column averages the six frameworks with equal weight. We weight benchmark frameworks rather than metrics because the count runs from 3 (PathoROB) to 39 (Patho-Bench), and a metric-weighted average would be decided almost entirely by Patho-Bench. The benchmark frameworks use different tasks and metrics, so the average rank summarizes relative standing across evaluations rather than placing the models on one scale of quality. HERO had the lowest average rank (2.61), followed by H-Optimus-0/1 (3.04), Virchow2 (3.18), and UNI2 (3.21), and ranked first on PathoROB, PLISM, and Patho-Bench. The slide counts on the horizontal axis of Fig.~\ref{fig:overall_ranking}b are those reported in each model's report.
\input{tables/S1_overall_ranking}

%% file: tables/07_ablation_summary.tex
\begin{table}[H]
\centering
\caption{\textbf{Effect of Stage~2 Gram anchoring on benchmark averages.} Stage~1 only is the DINO/iBOT checkpoint at 450k steps; + Gram anchoring continues it for 20k steps with the Gram-anchoring loss and is the released HERO model. $\Delta$: relative change from Stage~1 only to + Gram anchoring, \textcolor{deltaup}{$\uparrow$ green} for an increase and \textcolor{deltadown}{$\downarrow$ red} for a decrease. The released configuration is shaded.}
\label{tab:ablation_summary}
\small
\setlength{\tabcolsep}{9pt}
\renewcommand{\arraystretch}{1.15}
\begin{tabular}{lccc}
\toprule
Training recipe & \makecell{EVA\\{\scriptsize Average $\uparrow$}} & \makecell{HEST\\{\scriptsize Pearson $r\,\uparrow$}} & \makecell{PathoROB\\{\scriptsize RI $\uparrow$}} \\
\midrule
Stage~1 only (DINO/iBOT) & 0.776 & 0.414 & 0.881 \\
\rowcolor{heroblue}\textbf{+ Gram anchoring} & 0.780 & 0.404 & 0.892 \\
\addlinespace[2pt]
$\Delta$ (\%) & \up{0.5} & \down{2.4} & \up{1.2} \\
\bottomrule
\end{tabular}
\end{table}

%% file: tables/10_ablation_pathorob.tex
\begin{table}[H]
\centering
\caption{\textbf{PathoROB with and without Gram anchoring.} Robustness Index as in Table~\ref{tab:pathorob_results}. $\Delta$: relative change from Stage~1 only to + Gram anchoring, \textcolor{deltaup}{$\uparrow$ green} for an increase and \textcolor{deltadown}{$\downarrow$ red} for a decrease. The released configuration is shaded.}
\label{tab:ablation_pathorob}
\small
\setlength{\tabcolsep}{9pt}
\renewcommand{\arraystretch}{1.15}
\begin{tabular}{lcccc}
\toprule
Training recipe & \makecell{Camelyon\\{\scriptsize Breast}} & \makecell{TCGA 2$\times$2\\{\scriptsize Multi-organ}} & \makecell{Tolkach ESCA\\{\scriptsize Esophagus}} & \makecell{Average\\{\scriptsize RI $\uparrow$}} \\
\midrule
Stage~1 only & 0.813 & 0.873 & 0.956 & 0.881 \\
\rowcolor{heroblue}\textbf{+ Gram anchoring} & 0.836 & 0.884 & 0.956 & 0.892 \\
\addlinespace[2pt]
$\Delta$ (\%) & \up{2.8} & \up{1.3} & \same & \up{1.2} \\
\bottomrule
\end{tabular}
\end{table}

%% file: tables/08_ablation_eva.tex
\begin{table}[H]
\centering
\caption{\textbf{EVA with and without Gram anchoring.} Metrics as in Table~\ref{tab:eva_results}. $\Delta$: relative change from Stage~1 only to + Gram anchoring, \textcolor{deltaup}{$\uparrow$ green} for an increase and \textcolor{deltadown}{$\downarrow$ red} for a decrease. The released configuration is shaded.}
\label{tab:ablation_eva}
\scriptsize
\setlength{\tabcolsep}{2.7pt}
\renewcommand{\arraystretch}{1.12}
\resizebox{\linewidth}{!}{%
\begin{tabular}{lccccccccccc}
\toprule
\multirow{2}{*}{Training recipe}
& \multicolumn{8}{c}{Tile-level} & \multicolumn{2}{c}{Slide-level} & \multirow{2}{*}{Average} \\
\cmidrule(lr){2-9} \cmidrule(lr){10-11}
& PCam$_{10}$ & BACH & BRACS & CRC & PCam & Gleason & CoNSeP & MoNuSAC & Cam16 & PANDA & \\
\midrule
Stage~1 only & 0.866 & 0.832 & 0.629 & 0.967 & 0.945 & 0.776 & 0.633 & 0.644 & 0.838 & 0.633 & 0.776 \\
\rowcolor{heroblue}\textbf{+ Gram anchoring} & 0.875 & 0.848 & 0.622 & 0.963 & 0.943 & 0.793 & 0.633 & 0.657 & 0.834 & 0.636 & 0.780 \\
\addlinespace[2pt]
$\Delta$ (\%) & \up{1.0} & \up{1.9} & \down{1.1} & \down{0.4} & \down{0.2} & \up{2.2} & \same & \up{2.0} & \down{0.5} & \up{0.5} & \up{0.5} \\
\bottomrule
\end{tabular}%
}
\end{table}

%% file: tables/09_ablation_hest.tex
\begin{table}[H]
\centering
\caption{\textbf{HEST with and without Gram anchoring.} Metric as in Table~\ref{tab:hest_results}. $\Delta$: relative change from Stage~1 only to + Gram anchoring, \textcolor{deltaup}{$\uparrow$ green} for an increase and \textcolor{deltadown}{$\downarrow$ red} for a decrease. The released configuration is shaded.}
\label{tab:ablation_hest}
\scriptsize
\setlength{\tabcolsep}{3.0pt}
\renewcommand{\arraystretch}{1.12}
\resizebox{\linewidth}{!}{%
\begin{tabular}{lcccccccccc}
\toprule
Training recipe & \makecell{IDC\\{\scriptsize Breast}} & \makecell{PRAD\\{\scriptsize Prostate}} & \makecell{PAAD\\{\scriptsize Pancreas}} & \makecell{SKCM\\{\scriptsize Skin}} & \makecell{COAD\\{\scriptsize Colon}} & \makecell{READ\\{\scriptsize Rectum}} & \makecell{ccRCC\\{\scriptsize Kidney}} & \makecell{LUAD\\{\scriptsize Lung}} & \makecell{LYMPH-IDC\\{\scriptsize Lymph node}} & \makecell{Average\\{\scriptsize Pearson $r\,\uparrow$}} \\
\midrule
Stage~1 only & 0.572 & 0.374 & 0.477 & 0.666 & 0.303 & 0.216 & 0.293 & 0.562 & 0.261 & 0.414 \\
\rowcolor{heroblue}\textbf{+ Gram anchoring} & 0.585 & 0.378 & 0.489 & 0.614 & 0.261 & 0.209 & 0.271 & 0.561 & 0.269 & 0.404 \\
\addlinespace[2pt]
$\Delta$ (\%) & \up{2.3} & \up{1.1} & \up{2.5} & \down{7.8} & \down{13.9} & \down{3.2} & \down{7.5} & \down{0.2} & \up{3.1} & \down{2.4} \\
\bottomrule
\end{tabular}%
}
\end{table}

%% file: tables/S1_overall_ranking.tex
\begin{table}[H]
\centering
\caption{\textbf{Average rank across benchmark frameworks.} Models are ranked (1 = best; ties share the average rank) on every metric a framework reports, that is, each column of Tables~\ref{tab:pathorob_results}--\ref{tab:hest_results} and each of the 39 Patho-Bench tasks. Ranks are averaged within each framework, and the last column averages the six frameworks with equal weight. Metrics per framework: PathoROB 3, PLISM 4, EVA 10, THUNDER 6, HEST 9, Patho-Bench 39. H-Optimus-0 and H-Optimus-1 are pooled as one entry. Best in bold, second best underlined.}
\label{tab:overall_ranking}
\small
\setlength{\tabcolsep}{4.2pt}
\renewcommand{\arraystretch}{1.12}
\begin{tabular}{lcccccc|c}
\toprule
\multirow{2}{*}{Model} & \multicolumn{6}{c}{Average rank within each benchmark} & \multirow{2}{*}{\makecell{Average rank\\across frameworks}} \\
\cmidrule(lr){2-7}
 & PathoROB & PLISM & EVA & THUNDER & HEST & Patho-Bench & \\
\midrule
\rowcolor{heroblue}\textbf{HERO} & \best{1.00} & \best{1.00} & 3.35 & 4.00 & 3.44 & \best{2.86} & \best{2.61} \\
H-Optimus-0/1 & 2.67 & \second{2.25} & 3.50 & 4.08 & \best{1.61} & 4.10 & \second{3.04} \\
Virchow2 & \second{2.33} & 2.75 & \second{2.70} & \second{3.08} & 3.67 & 4.56 & 3.18 \\
UNI2 & 4.00 & 5.25 & \best{2.40} & \best{2.00} & \second{2.44} & \second{3.18} & 3.21 \\
UNI & 5.33 & 5.50 & 4.90 & 4.00 & 5.39 & 3.72 & 4.81 \\
Prov-GigaPath & 6.00 & 5.00 & 4.55 & 5.00 & 5.22 & 4.45 & 5.04 \\
Phikon-v2 & 6.67 & 6.25 & 6.60 & 5.83 & 6.22 & 5.13 & 6.12 \\
\bottomrule
\end{tabular}
\end{table}